\documentclass[letterpaper, 10 pt, conference]{ieeeconf}  

\usepackage{graphics} 
\usepackage{graphicx}
\usepackage{hyperref}
\usepackage{cite}
\usepackage{mathtools}
\usepackage[nolist,printonlyused]{acronym}
\usepackage{cleveref}
\usepackage{multirow}
\usepackage{todonotes}

\title{\LARGE \bf
	SKILL-IL: Disentangling Skill and Knowledge in Multitask Imitation Learning
}

\author{Bian Xihan and Oscar Mendez and Simon Hadfield}

\begin{document}

\maketitle
\thispagestyle{empty}
\pagestyle{empty}

\begin{abstract}

In this work, we introduce a new perspective for learning transferable content in multi-task imitation learning. Humans are capable of transferring skills and knowledge. If we can cycle to work and drive to the store, we can also cycle to the store and drive to work. We take inspiration from this and hypothesize the latent memory of a policy network can be disentangled into two partitions. These contain either the knowledge of the environmental context for the task or the generalisable skill needed to solve the task. 
This allows an improved training efficiency and better generalization over previously unseen combinations of skills in the same environment, and the same task in unseen environments.

We used the proposed approach to train a disentangled agent for two different multi-task IL environments. In both cases, we out-performed the SOTA by 30\% in task success rate. We also demonstrated this for navigation on a real robot.

\end{abstract}

\section{INTRODUCTION}
\vspace{-1mm}
Multi-task learning is a fast-growing field in machine learning.
The essence of multi-task learning is to allow an agent to perform multiple different tasks without retraining.
This is often considered to be the most feasible route to the development of a general AI. However, facilitating multi-task learning in weakly supervised scenarios such as Imitation Learning (IL) is an emerging field. 
Imitation Learning operates through trial and error, but we can never perform enough trials to explore every possible combination of tasks and solutions. 

We must learn to generalize and share information across different domains and re-combine this information for unseen tasks.
Researchers struggle to transfer expertise efficiently between tasks, or even between sub-problems of the same task. Meanwhile, research on transferring to previously unseen tasks (zero-shot RL/IL) growing in popularity.
To approach this problem, we find inspiration from human learning behaviours.
We as humans spend years learning varied tasks, from how to walk and talk, to writing papers or gymnastics. 
This learning is a process of imprinting memories in our brain, not entirely dissimilar to training the weights of a neural network.
Procedural Memory, or ``\textbf{Skill}" is the memory required for the agent to perform a certain task in general \cite{mccormick1997conceptual,rittle1998relation}.
Declarative Memory, or ``\textbf{Knowledge}", involves memory specific to the environment the agent is operating in \cite{mccormick1997conceptual,burgin2016theory}.
For example, when we are driving to work, this requires us to have the \textbf{skill} of driving a car (procedural memories), and the \textbf{knowledge} of the route to get to work (declarative memories).
Most tasks require both skill and knowledge simultaneously to complete. However, these are independent and transferable. We can also use our driving skills to drive to the store, or we can use our knowledge of our workplace to cycle to work. Neither of these transfers would imply additional training.
This capability would be invaluable in multi-task learning, as each problem requires a different combination of knowledge and skill.
Generally, every possible combination of knowledge and skill is treated as a separate learning problem, or every skill is trained independently to generalize overall knowledge. This greatly increases the difficulty of multi-task learning, leading to scalability issues and unrealistic training requirements. 
In this paper, we propose Skill and Knowledge Independent Latent Learning (SKILL) as a new approach to multi-task IL which explicitly disentangles and shares both skills and knowledge across tasks, as shown in figure 1.
\begin{figure}
	\centering
	\includegraphics[width=0.9\linewidth]{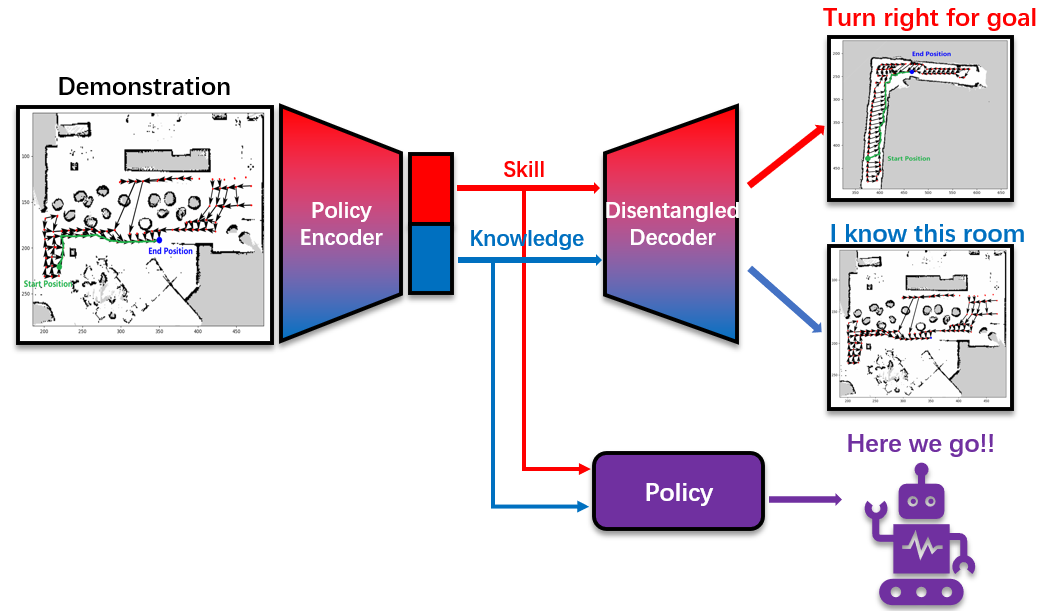}
	\caption{The policy encoder provides an embedding consisting of both skill and knowledge, coupled with the disentangled decoder to form a gated VAE architecture which partitions the embedded latent.}
	\label{fig:firstpageimg}
	\vspace{-5mm}
\end{figure}


In order to disentangle the learning of skill and knowledge, we need to adapt how we present training examples to the agent, as well as the architecture of the model.
Knowing that all tasks can be represented as a combination of skill and knowledge, we take inspiration from recent work on disentangled VAEs \cite{vowels2020gated} to learn a joint latent representation across all the tasks to be performed. 
This latent representation is partitioned into two subdomains dedicated to representing the skill and the knowledge of the task.
These latent subdomains are jointly trained in a weakly supervised manner, in parallel to learning a policy from the latent observation space.

We show experimentally that we can successfully disentangle the skill and knowledge in multi-task learning, Furthermore, we show that this improves training efficiency and final performance. To summarise, the main contributions of this study are as follows:

\begin{itemize}
	\item A self-supervised VAE-based architecture to learn a disentangled representation of robotic tasks
	\item A multi-task imitation Learning approach which shares training experiences across latent subdomains
	\item An approach to generate a more human-interpretable latent space for multi-task imitation learning, enabling decoding and visualization of the latent for better understanding.

\end{itemize}

\vspace{-2mm}
\section{Literature Review}
\vspace{-0.5mm}

\subsection{Disentangled Representations}
\vspace{-0.5mm}
The state of the art for learning disentangled representations is dominated by VAE approaches.
VAEs, or variational auto-encoders, are generative models which re-parametrize a latent space as a distribution to be sampled from. 
Each dimension of the latent representation learned by the VAE is generally considered to be an independent generative factor \cite{dai2019diagnosing}.
These elements in the representation can capture and isolate certain underlying factors without affecting other elements in the latent space.
A great deal of research has been done to further explore the disentanglement of these learned representations\cite{zheng2019disentangling,ansari2019hyperprior}. 
In beta-VAE and the later work of Burgess et al.\cite{higgins2016beta}, a variation of the VAE framework is proposed which balances the latent channel capacity and constraints with reconstruction accuracy. 
The work of Vowels et al.\cite{vowels2020gated} overturned this paradigm through a weakly-supervised approach which isolates domain knowledge in the training process of a gated VAE. 
This framework makes it possible to learn latent subdomains, by appropriately partitioning the training based on shared properties. The learning of the latent factors is still unsupervised, but additional losses are provided as a soft constraint to group the factors into subdomains.
This method was shown to be more informative and has a better quality of disentanglement.
We take inspiration from this and propose an algorithm to learn latent skill and knowledge subdomains.

\vspace{-1.5mm}
\subsection{Multi-Task Learning}
\vspace{-0.5mm}

In hierarchical multi-task research, sub-tasks are often learned through linguistically categorised representations of a specific set of tasks. 
The representations used by these systems sometimes unintentionally explore the skill/knowledge paradigm.
The work of Oh et al.\cite{oh2017zero}, introduced the innovative ``analogy" representation of subtasks.
Here the target objects and the actions which can be applied to a target object are independent. 
The work of Bian et al. \cite{bian2021robot} addressed a single type of task but focussed on learning different behaviours in different types of environments.
The work of \cite{devin2019compositional} introduced Compositional Plan Vectors (CPV), which instead of learning the representation of each subtask, the network learns the embedding for a composition of a sequence of sub-tasks. This allows the decomposition of tasks without hierarchical or relational supervision.
Our work bridges these different ideas, learning a latent space where not only can subtasks be composited, but where the skill and knowledge components of subtasks can be shared and permuted. This makes it possible to solve never before seen task combinations, as a step towards zero-shot IL and general AI.

\vspace{-2mm}
\section{Methodology}
\vspace{-1mm}
\begin{figure}
	\centering
	\includegraphics[width=0.75\linewidth]{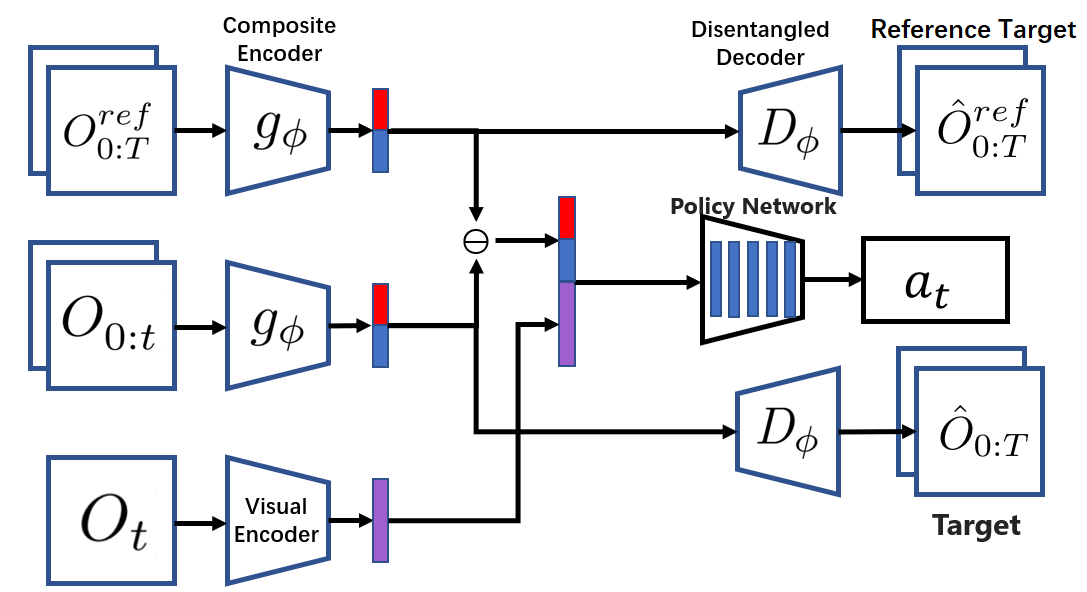}
	\caption{The network requires a current state $O_t$, a reference trajectory $O^{ref}_{0:T}$, and the current trajectory $O_{0:t}$. The output consists of both an action and two reconstructed image pairs for the reference input and the current input.}
	\label{fig:archfig2}
	\vspace{-6mm}
\end{figure}

In this paper, we introduce the Skill and Knowledge Independent Latent Learning (SKILL) architecture, as shown in Figure \ref{fig:archfig2}. 
With this architecture and training framework, we attempt to disentangle the learned skill and knowledge within a latent embedding, and divide the latent into two subdomains each containing the learned skill and knowledge.
This architecture consists of a pair of gated VAEs \cite{vowels2020gated} that partition the latent space. Each subdomain includes a masked latent and is updated by similarity loss between pairs of examples.
The gated VAEs are given two sets of trajectories which consist of the start and end state as well as the current point of the agent. These are used to produce a CPV which plans the action from the current state to the end state.
The gated VAEs are trained using pairs of experiences which share either the same environment, the same task list, or both, to reinforce the learning of the corresponding subdomain. Similar to humans learning how to drive a car by driving to different places, and learning the city's layout by travelling around the city with different modes of transport.

The agent is required to perform tasks, which include modifying its environment to reach a target goal state. The agent may need to complete several subtasks in order to complete a task. 
The sequence of states visited during the completion of the task is referred to as the trajectory of the agent.
The input does not specify any particular ordering of the subtasks within the trajectory. 
The set of subtasks is implicit, which gives the agent freedom to determine the best set of subtasks to reach a particular goal state.

The embedding of the portion of the task remaining to complete (current-to-end) is calculated as the difference between the embedding of the overall task (start-to-end), and the embedding of the progress so far (start-to-current). 
This is combined with a visual embedding and passed to the policy network to determine the agent's following action.

In our experiments we perform imitation learning, taking the full set of states from timestep $t=0$ to the final timestep $t=T$, let $O$ be the observation of a state in a fully observable environment, then $O_0^{ref}...O_T^{ref}$ is an expert reference trajectory. 
The expert reference trajectories are extracted by a greedy search over the environment for the optimal solution.


\vspace{-1.5mm}
\subsection{Variational Task Embedding}
\vspace{-1mm}

We will first define a compositional representation for any combination of sub-tasks in a multi-task learning environment to acquire a latent that embeds both skill and knowledge.
A compositional representation is an embedding which encodes structural relationships between the items in the space \cite{mikolov2013distributed}. 
This representation which contains both skill and knowledge will be used as the full latent where the disentanglement takes place.
The multi-task environment will provide the disentanglement by allowing tasks and environments to be mixed in different combinations.
Consider a compositional task embedding $\vec{v}$ which encodes a set of tasks as the sum of the compositional embeddings for all subtasks. 
To avoid enforcing a particular ordering for the completion of these subtasks, our planning space is built with commutativity, i.e. A+B = B+A.
Given this definition, the embedding of all tasks that have yet to be accomplished can be calculated as $(\vec{v}-\vec{u})$ where $\vec{u}$ is the embedding vector for the tasks accomplished so far.
As we focus on semi-supervised machine learning, we don't specify the exact end state for the agent.
Instead, the policy $\pi(a_t|O_{t},\vec{v}-\vec{u})$ produces the action $a_t$ based on the current state $O_{t}$ and the ``to do" task embedding. 

Next, we introduce the different losses for the model.
Let function $g_{\phi}(O_{a:b})$ encode the observation pair at time $a$ and $b$ into a latent task embedding using parameters $\phi$. 
To help further the learning of the task embedding, the function $g$ is a probabilistic encoder which predicts means and variances for each latent parameter. This is coupled with a decoder $d_\psi$ to form a VAE, such that $O \approx d_\psi(\vec{u})$ where $\vec{u} \sim g_\phi(O)$. 
We define the reconstruction error against target $ \hat{O} $ as $l_{rec}(O, \hat{O} )=|d_\psi(g_{\phi}(O) -  \hat{O} )|$ where the intermediate sampling step is omitted for brevity. The full reconstruction loss is obtained by applying this to both the reference trajectory $ (O_{0:T}^{ref})$ and current trajectory $ (O_{0:t})$ inputs 
\vspace{-1mm}
\begin{equation}
L_{\delta}(O^{ref}_{0:T},O_{0:t},\hat{O}^{ref},\hat{O}) = l_{\delta}\left(O_{0:T}^{ref},  \hat{O} ^{ref}\right)+l_{\delta}\left(O_{0:t},  \hat{O} \right)
\vspace{-1mm}
\end{equation}
To reduce the impact of empty space, we also mask the reconstruction loss to only include non-zero pixels.

During the forward pass, as both skill and knowledge are required to solve the task, the entire latent space is used by the policy network to select an action. Therefore, the policy function is:
$
\pi \left(a_t|O_t, g_{\phi}\left(O_{0:T}^{ref}\right)-g_{\phi}\left( O_{0:t}\right)\right)
$. 
Hence the policy loss $L_{a}$ is given by the loss function:
\vspace{-1mm}
\begin{equation}
L_{a}(O_{t}, \phi )\! = \!- \!log\!\left(\! \pi\!\left(\!\hat{a}_t|O_{t},g_{\phi}\!\left(\!O_{0:T}^{ref}\right)\!-\!g_{\phi}\!\left( \!O_{0:t}\!\right)\!\right)\!\right)
\vspace{-0.5mm}
\end{equation}
where $\hat{a}_t$ is the reference action.

Additionally, there are two regularization losses using the triplet margin loss $l_m$ from \cite{schroff2015facenet}. The first $L_C$ enforces the compositionality of the latent space by ensuring that the sum of the embeddings for partial completion ($u_{0:t}$) and the embedded to-do vector ($u_{t:T}$) are equal to the embedding for the entire task ($u_{0:T}$). 
\vspace{-2mm}
\begin{equation}
L_C(O_0, O_t, O_T) = l_m(g(O_{0:t}) + g(O^{ref}_{t:T}) - g(O^{ref}_{0:T}))
\vspace{-0.5mm}
\end{equation}
where $l_m$ is a truncated L1 loss with the margin equal to 1.
The second regularization loss tries to ensure that similarity in the latent space corresponds to semantically similar tasks. To this end, we ensure that the embedding of our agent's trajectory is similar to that of the embedding of the expert's reference trajectory
\vspace{-3mm}
\begin{equation}
L_P = l_m(g(O_{0:T})-g(O^{ref}_{0:T}))
\vspace{-1mm}
\end{equation}
The sum of these two loss functions is used to regularize the model: $L_{R} = L_C + L_P$.
The latent representation used by the agent comprises both the ability to solve the current task, which is the skill; and the information about the current environment, which is the knowledge.
However, these two types of latent information are currently entangled.



\vspace{-0.5mm}
\subsection{Disentangling Skill and Knowledge Subdomains} 
\vspace{-0.5mm}
To disentangle the task vectors ($\vec{u}$) into skill and knowledge sub-domains($\vec{u}= [\vec{u}^s, \vec{u}^k]$), we utilize the gated VAE \cite{vowels2020gated} approach with the CPV encoders as part of the VAE. 
The input and target images are first grouped according to shared skill factors or shared knowledge factors. 
Specifically, if two training examples ($O$ and $ \hat{O} $) both comprise the same sequence of subtasks but within a different environment, these examples are grouped by skill and added to the skill training set $\mathcal{S} = \mathcal{S} \cup (O,  \hat{O} )$.
Similarly, if the training examples comprise different sequences of subtasks, but within the same environment, they are grouped by knowledge and added to the knowledge training set $\mathcal{K} = \mathcal{K} \cup (O,  \hat{O} )$. 
In this work we enforce a hard gating by partitioning the latent space into two non-overlapping regions, the ratio of the sizes of these two latent subdomains can be changed based on the task. In all our experiments we kept them equal, each representing either skill or knowledge. 

To disentangle the skill from knowledge, we adapt the reconstruction loss from equation 1. The input and target pair for both terms are drawn from either the skill or knowledge training set such that $(O,  \hat{O} ) \in (\mathcal{S} \cup \mathcal{K})$. We additionally adapt which partition of the latent space is updated via backpropagation based on this.

More formally, we define $\lfloor\rfloor$ as an operator which masks gradients during the back pass. We then define the gated latent space as
\vspace{-2.5mm}
\begin{equation}
\vec{u} =\begin{cases} 
	[\vec{u}^s, \lfloor \vec{u}^k \rfloor] &if (O,  \hat{O} ) \in \mathcal{S} \\
	[\lfloor \vec{u}^s \rfloor, \vec{u}^k] &if (O,  \hat{O} ) \in \mathcal{K} \\
	[\vec{u}^s, \vec{u}^k] &if (O,  \hat{O} ) \in \mathcal{S} \cap \mathcal{K} \\
	\end{cases}
\vspace{-0.5mm}
\end{equation}
This means that for each training pair, gradient flow and parameter updates only occur for the subdivision of the latent space which is shared by the source and the target. For further details on gated VAEs, we refer the reader to \cite{vowels2020gated}. In summary, this approach makes it possible to learn disentangled latent subdomains without knowing the ground truth latent factors. We only need to be able to cluster examples based on shared subdomains.
It is worth noting that within our framework the grouping and subsequent selection of skill or knowledge targets is done for both the current branch $(O,\hat{O})$ and the reference branch $(O^{ref}, \hat{O}^{ref})$.

Additionally, we introduce a dynamic loss $L_G$.
While $\alpha, \beta$ are regularization constants, it is possible to improve the disentanglement performance by changing the value of $\alpha$ and $\beta$ according to the training mode. This can be expressed as:
\vspace{-3mm}
\begin{equation}
L_G = 
\begin{cases}
	\epsilon\alpha L_{a} + \beta L_{\delta}  & if (O,  \hat{O} ) \in \mathcal{S} \\
	\alpha L_{a} + \epsilon\beta L_{\delta} & if (O,  \hat{O} ) \in \mathcal{K}\\
	\alpha L_{a} + \beta L_{\delta}  & if(O,\hat{O}) \in S \cap K
\end{cases}
\label{eq8}
\vspace{-0.5mm}
\end{equation} 
where $\epsilon$ is a small value constant.

The reason behind this dynamic loss weighting is we expect the agent to correctly predict the policy action during skill training. However, as the environment is different from the original episode, the reconstruction loss is expected to be less important. Similarly, the $\alpha$ value can be lowered to ensure the reconstruction loss is emphasised during knowledge training.

To summarize, the loss function $L$ of the framework comprises both reconstruction loss and policy loss with the dynamic loss weighting, summed with the regularization loss: $L = L_G + L_{R}$.

\section{Evaluation}
\vspace{-0.5mm}
We evaluate the SKILL framework to show how the proposed disentanglement of skill and knowledge impacts both the agent's success rate and efficiency. We perform a range of qualitative experiments, exploring and confirming the level of disentanglement learned by our system. Following this, we explore the importance of different elements of our system via an ablation study. We also evaluate this across two different environments and compare it against the current state-of-the-art technique in each. Finally, we demonstrate our technique with a real robot performing navigation tasks.

\paragraph{Craftworld Environment}
The first environment used in our experiments is a Minecraft-inspired 2D crafting world \cite{devin_craftingworld_2020}. 
The world has a discrete state and action. The agent is allowed to move, pick up or drop off certain items present on the map, as well as perform actions on those items. 
With this environment, we can define tasks such as chop trees, break rock, make bread, build a house, etc. and combine them into sequences such as [make bread, eat bread, chop trees, build house]. This provides a good selection of unique tasks and sequences to generate training data.
As detailed in the methodology section, the objectives of the agent are specified implicitly by providing two trajectories consisting of observations of both the current trajectory and the reference trajectory. 
The advantage of this approach is that no explicit ordering of subtasks is specified, and the agent is free to execute tasks in the most appropriate manner. 
Our framework is trained with randomly generated starting environments and random combinations of tasks to complete. 
The complexity of the problem increases as more tasks are required to reach the target end state. The previous state-of-the-art approach in this environment \cite{devin2019compositional} used the same input observations and expert reference trajectories.
\begin{figure}
	\centering
	\includegraphics[width=0.8\linewidth]{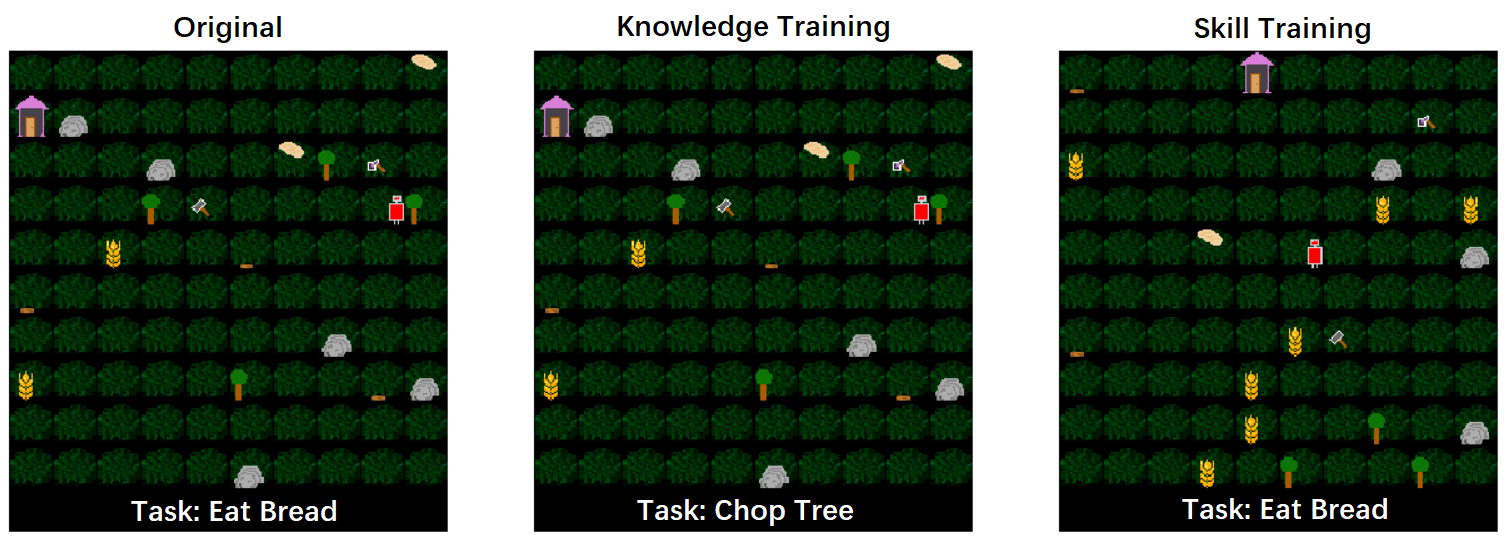}
	\caption{The different inputs for different training modes. In skill mode, the environment differs from the original but the agent is expected to perform the same task. In knowledge mode, the environment is the same but the agent is expected to perform a different task.}
	\label{fig:firstenv}
	\vspace{-5mm}
\end{figure}

The model is given three sets of data as shown in figure 3. Firstly, an original episode with an environment and a sequence of tasks. Secondly, an episode with the same environment but different tasks for knowledge training. Thirdly, an episode with the same tasks and a different environment for skill training.
In figure 3, the \textit{original} episode requires the agent to pick up a hammer and break a rock. In the \textit{knowledge training} episode, the environment is the same, but the task is to pick up wheat and make bread. In the \textit{skill training} episode, the environment is different from the original episode, but the task is once again to use a hammer to break the rock.
\paragraph{Learned Navigation}
The second environment simulates a 2D navigation scenario. The maps are created from gmapping \cite{grisettiyz2005improving} outputs in real-world locations to simulate real-world navigation as shown in figure 4. 
The goal in this environment is to reach a random target location on the map. The agent is given a full state observation as well as a demonstration during training.  
The state-of-the-art (SOTA)\cite{bian2021robot} technique treated this multi-environment navigation problem as a multi-task learning scenario, using the camera view rather than the map as input. 
Nevertheless, the action space and the quantized state space remain the same as in the original paper.
In both environments, we focus on two evaluation metrics: the task success rate measures how many episodes end in the goal being successfully reached. The average episode length measures how quickly the agent was able to achieve its goal.
\begin{figure}
	\centering
	\includegraphics[width=0.8\linewidth]{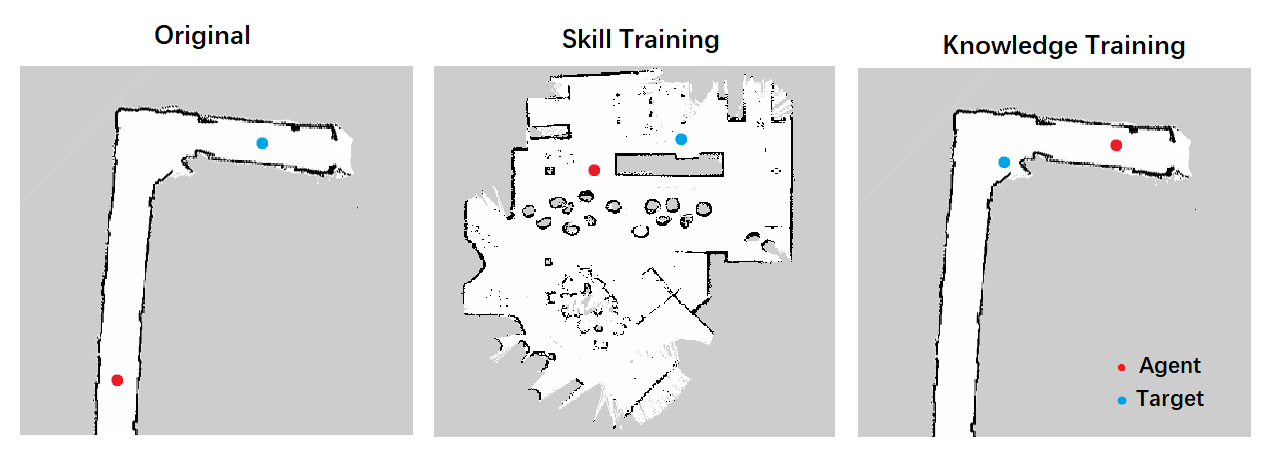}
	\vspace{-1.5mm}
	\caption{The navigation environment mimics real maps produced by the gmapping \cite{grisettiyz2005improving} algorithm.}
	\label{fig:secondenv}
	\vspace{-6mm}
\end{figure}

\vspace{-2mm}
\subsection{Implementation}
\vspace{-1mm}
In both environments, the observation is provided as a pair of images. 
The encoder $g$ shared by the reference trajectories $O_{0:T}^{ref}$ and the current trajectories $O_{0:t}$ is a 4-layer CNN encoder with shared weights. 
The current state input ($O_t$) is processed by a 4-layer CNN with a final fully connected layer.
The latent embedding of the task left to complete is fed into a 5-layer policy network to produce the action during each time step, while each disentangled latent is fed into a 4-layer decoder for reconstruction.
\vspace{-1mm}
\subsection{Exploring Disentanglement}
\vspace{-1mm}
In our first set of experiments, we seek to confirm whether our proposed approach results in a latent space where skill and knowledge are disentangled. Unfortunately, measuring disentanglement is extremely challenging. There are many proposed approaches in the literature but most require the ground-truth factors to be known. 
Instead, to quantify our disentanglement of skill and knowledge, we first take our trained model and freeze the network weights.
Next, we record the latent embeddings produced by our network for all samples in the dataset.
Finally, we attempt to train a simple network that estimates the task id from only one of the latent partitions ($\vec{u}^s$ or $\vec{u}^k$).

When testing on a held-out set of 500 unseen latent embeddings, the network trained on the skill partition achieved a 99.2\% accuracy in recovering 6 different task labels. However, for the network trained on the knowledge latent, the recovery accuracy is only 12.7\%.
This shows that all the information relating to the skill which solves each specific task has been effectively disentangled and concentrated into the latent skill subdomain.

We could not perform a similar test for the knowledge subdomain because we do not have a fixed number of environmental layouts to recognise. Instead, we trained two image decoder networks which attempt to reconstruct the environment using only the skill or knowledge latent partitions respectively.
The decoder networks use the latent partitions as input to generate the corresponding observations, both are trained until they reach their peak accuracy.
Examples of reconstructed images for previously unseen latent embeddings are shown in figure 5.
\vspace{-1.5mm}
\begin{figure}[tbh!]
	\centering
	\includegraphics[width=0.7\linewidth]{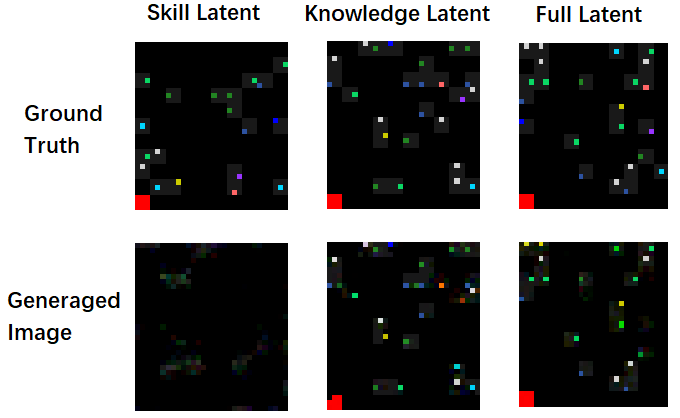}
	\vspace{-1.5mm}
	\caption{
	The reconstructed image from the knowledge latent recreated the original image almost perfectly, full latent recreated the image without items unrelated to the current task (red hammer and purple house are not related to chop trees), and skill latent fails to generate an image that resembles the ground truth.}
	\label{fig:exp1}
	\vspace{-3mm}
\end{figure}

It is apparent that the reconstruction results using the previously unseen knowledge latent are much better than the results from the skill latent. We note that the skill latent alone is unable to produce any meaningful image, as it contains little environmental information. Meanwhile, the knowledge-only reconstructions appear to focus on representing the most salient parts of the environment. The reconstruction from the full latent can reconstruct the environment without some unessential items not used in the current task. This shows that some of the less salient final details may be jointly encoded across both latent subdomains. 
This mirrors findings in neuroscience which indicate that in biological learning, declarative and procedural knowledge can be disentangled to a great extent but never completely.

Numerically, the average reconstruction loss across the validation dataset for the knowledge latent is around 300 times lower than the reconstruction loss from the skill latent.
We find similar results in the second environment, where the average reconstruction loss across the knowledge latent dataset is around 250 times lower than with the skill latent.

\begin{table}[tbh]
	\centering
	\resizebox{0.45\textwidth}{!}{%
		\begin{tabular}{llll}
			Model &
			\begin{tabular}[c]{@{}l@{}}Imitation\\ accuracy\end{tabular} &
			\begin{tabular}[c]{@{}l@{}}Success\end{tabular} &
			\begin{tabular}[c]{@{}l@{}}Ep. length\end{tabular} \\ \hline
			CPV-FULL\cite{devin2019compositional}       & 66.42\%          & 65\%          & 69.31    \\
			SKILL-no $O_t$ & 64.18\%          & 65\%          & 26.95          \\
			SKILL   & 70.61\%          & 84\%          & 19.77          \\
			SKILL+FS    & \textbf{70.89\%} & 89\%          & 19.52          \\
			SKILL+FS+DL   & 70.62\%          & \textbf{94\%} & \textbf{17.88}
			     
		\end{tabular}%
	}
	\vspace{-2mm}
	\caption{\vspace{-2mm}Ablation study. FS indicates fixed sampling. DL indicates dynamic loss weighting.}
	\label{tab:my-table}
	\vspace{-1mm}
\end{table}

\vspace{-1mm}
\subsection{Ablation Study}

\begin{table}
	\vspace{-5mm}
	\centering
	\resizebox{0.4\textwidth}{!}{%
		\begin{tabular}{lll}
			Model       & \begin{tabular}[c]{@{}l@{}} Success\end{tabular} & Ep. Length \\ \hline
			SKILL+FS    & 90\%                                                    & 14.82      \\
			SKILL+FS+DL & 96\%                                                    & 13.08      \\
			SKILL+FS+KL & \textbf{98\%}                                                    & 13.31      \\
			SKILL+FS+SL & 84\%                                                    & \textbf{11.47}      
		\end{tabular}%
	}\vspace{-1mm}
	\caption{\vspace{-2mm}Ablation study for Env.2. KL indicates higher knowledge partition, SL indicates higher skill partition.}
	\vspace{-9mm}
	\label{tab:my-table2}
\end{table}
\begin{table*}[th]
	\centering
	
	\resizebox{\textwidth}{!}{%
		\begin{tabular}{lllllll|llllll}
			\hline
			\multicolumn{1}{l|}{\multirow{2}{*}{MODEL}} &
			\multicolumn{2}{l|}{4 SKILLS} &
			\multicolumn{2}{l|}{8 SKILLS} &
			\multicolumn{2}{l|}{16 SKILLS} &
			\multicolumn{2}{l|}{1,1} &
			\multicolumn{2}{l|}{2,2} &
			\multicolumn{2}{l}{4,4} \\
			\multicolumn{1}{l|}{} &
			Success &
			\multicolumn{1}{l|}{Ep. Length} &
			Success &
			\multicolumn{1}{l|}{Ep. Length} &
			Success &
			Ep. Length &
			Success &
			\multicolumn{1}{l|}{Ep. Length} &
			Success &
			\multicolumn{1}{l|}{Ep. Length} &
			Success &
			Ep. Length \\ \hline
			CPV-NAÏVE\cite{devin2019compositional} &
			52.5 & 82.3 & 29.4 & 157.9 & 17.5 & 328.9 & 
			57.7 & \textbf{36.0} & 0.0 & -- & 0.0 & -- \\
			CPV-FULL\cite{devin2019compositional} &
			\textbf{71.8} & 83.3 & 37.3 & 142.8 & \textbf{22.0} & 295.8 &
			73.0 & 69.3 & \textbf{58.0} & 270.2 & 20.0 & 379.8 \\ \hline
			SKILL &
			61.3 & \textbf{63.3} & \textbf{37.5} & \textbf{132.7} & 20.0 & \textbf{277.8} &
			\textbf{80.0} & 53.3 & 55.0 & \textbf{103.1} & \textbf{26.3} & \textbf{198.1}
		\end{tabular}%
	}
\vspace{-4mm}
	\caption{Comparing Against SOTA in the craft world environment}
	\label{tab:my-table3}
	\vspace{-9mm}
\end{table*}
Now that we have conclusively demonstrated the successful disentanglement of our learned embedding space, we next perform an ablation analysis of our system. To this end, we explore the contributions of 3 parts of our model. For this experiment, we additionally report the imitation accuracy (percentage of actions that agree with the expert) for comparison \cite{devin2019compositional}. 
As shown in table \ref{tab:my-table}, we first removed the current state observation branch (no $O_t$). With only the gated VAE structure, our model performs similarly to the SOTA model (CPV-FULL\cite{devin2019compositional}). 
Introducing the current state observation branch improves performance significantly by giving the agent a more direct observation of its current state.
We then remove the random sampling from the latent distribution, and instead simply take the mean latent embedding. We refer to this as Fixed Sampling (FS).
This offers a small improvement in all metrics.
Finally, we introduce the dynamic loss (DL) weighting scheme proposed in equation \ref{eq8}. This approach provides further improvement in task performance and completion speed. 
Adjusting the proportions of the loss functions according to this training mode will improve training stability at the cost of reducing the training speed, as the learning happens less aggressively.
It is interesting to note that despite imitation accuracy peaked without the DL model, task success and completion speed still have improved through applying the DL model. 

In the second environment, we study the effect of different partition ratios shown in table \ref{tab:my-table2}. We used FS and FS+DL as references, with the latent space split evenly between skill and knowledge.
When we allocate more of the latent space to the knowledge subdomain (KL), the result surpasses the FS+DL model in both success rate and speed. 
When a higher partition is given to the skill subdomain (SL), while the task success rate dropped by 15\%, the completion speed increased significantly. 
This indicates an interesting trade-off between environmental knowledge for successful navigation and skill for efficiency.

\vspace{-1mm}
\subsection{Comparison vs State-Of-The-Art}
\vspace{-1mm}
After determining the optimal approach, we will now compare our model more thoroughly against the previous SOTA model (CPV-FULL\cite{devin2019compositional}) in both environments. For craftworld \cite{devin_craftingworld_2020}, we follow the evaluation protocol in \cite{devin2019compositional}. Both our model and the SOTA model are trained on 50,000 samples from sequences with 1-3 different tasks, and we evaluate each model against sequences with 4,8, and 16 tasks. We also evaluated the model's capability with a sequence of tasks with ``1,1" being a single task, and ``2,2" being a sequence of 2 tasks from 2 reference trajectories. As shown in table \ref{tab:my-table3}, our model outperforms the SOTA model in both task success rate as well as performance speed in most cases. In particular, our technique leads to a 30\% relative increase in the success rate of the most challenging experiment, and a 50\% reduction in episode length. This indicates that our model has a better generalization capability when dealing with trajectories with more tasks, as well as when dealing with a composing sequence of tasks. In the navigation environment (figure \ref{fig:secondenv}), 
the previous SOTA \cite{bian2021robot} has a success rate of 94.6\%. 
while our model can achieve a 98.0\% success rate. The average efficiency of our agent is also 20\%-30\% faster than then the previous SOTA. 
With the disentanglement of skill and knowledge, we can better share useful experiences across different navigation tasks.

\vspace{-1mm}
\subsection{Real Life Demonstration}
\vspace{-1mm}
Lastly, we demonstrate our model with a live turtlebot3\cite{amsters2019turtlebot} as the platform. 
The turtlebot first creates a map of the area using gmapping\cite{grisettiyz2005improving}, which is then processed into an observation format recognizable by the agent. The target location is marked on the observation along with the robot's current location.
The agent will produce a command for the robot to go in one of four directions for a set length.
This process repeats until the robot reaches the target location.
Without any fine-tuning, the robot learned to compensate for odometry inaccuracies and drift by performing recovery moves during the navigation, even though it was not exposed to this drift during the simulated training.
Our robot can successfully complete the task in multiple locations, a screen capture of the demonstration video is shown in the video attachment. 
\vspace{-4mm}
\section{CONCLUSIONS}
\vspace{-1mm}
In this work, we approached the problem of multi-task learning from a new perspective. 
Taking inspiration from neurobiology and pedagogy on memory acquisition, we hypothesized the latent space in a policy neural network could be disentangled into subdomains. 
Each partition is responsible for either the skill or the knowledge of the task and should be transferable to different combinations of future experiences. 
We successfully demonstrated this disentanglement in imitation learning, using a gated VAE architecture. 
With our method, we out-perform the SOTA model in two different environments, both in terms of success rate and speed.

To be able to disentangle the skill and knowledge in a task is a fundamental step toward combinational generalization. 
A better model to partition the skill and knowledge latent or to explain the entangled information will benefit our understanding of imitation learning in general.
Human interpretable solutions to complex tasks are also an interesting direction as it's been a popular choice in multi-task learning.
\looseness=-1

\addtolength{\textheight}{-9cm}   



%
%

\section*{ACKNOWLEDGMENT}

	This work was partially supported by the UK Engineering and Physical Sciences Research Council (EPSRC) grant agreement EP/S035761/1 "Reflexive Robotics".
\vspace{-1mm}
%


\bibliographystyle{plain}
\bibliography{cpVAE.bib}

\end{document}